\documentclass[letterpaper, 10 pt, conference]{ieeeconf}  

\IEEEoverridecommandlockouts                              

\usepackage{svg}
\usepackage{hyperref}
\usepackage{amsmath} 
\usepackage{amssymb} 
\usepackage{import}
\usepackage{color}
\usepackage{graphicx}
\usepackage{booktabs}
\usepackage{pgf} 
\usepackage{siunitx}
\usepackage{tabularx}
\usepackage[outline]{contour}
\contourlength{0.4pt}

\newcommand{\cspi}{\textsc{CSP-iNat}}
\newcommand{\cspf}{\textsc{CSP-FMOW}}
\newcommand{\ident}{\textsc{Identity}}
\newcommand{\range}{\textsc{RANGE}}

\newcommand{\melt}[1]{\MakeLowercase{\textsc{melt#1}}}
\newcommand{\salt}[1]{\MakeLowercase{\textsc{salt#1}}}
\newcommand{\textclip}{\textsc{TextCLIP}}
\newcommand{\natclip}{\textsc{NatCLIP}}

\title{\LARGE \bf Multi-Modal Contrastive Learning for Implicit Earth Embeddings\\ via Location Tying}

\author{Jonathan Hecht$^{1,3}$, Lukas Arzoumanidis$^{1}$, Ziyue Li$^{2}$ and Youness Dehbi$^{1}$
\thanks{$^{1}$Computational Methods Lab, HafenCity University Hamburg, Germany.
        {\tt\small \{lukas.arzoumanidis, youness.dehbi\}@hcu-hamburg.de}}%
\thanks{$^{2}$Dept. of Operations \& Technology, Technical University of Munich; Heilbronn Data Science Center; Munich Data Science Institute.
        {\tt\small ziyue.li@tum.de}}%
\thanks{$^{3}$Institute of Geodesy and Geoinformation, University of Bonn, Germany. {\tt\small jhecht@uni-bonn.de}}%
}

\begin{document}
\maketitle
\thispagestyle{empty}
\pagestyle{empty}

\begin{abstract}
Spatial prediction tasks are often limited by a lack of high-quality labelled ground-truth observations. To overcome this challenge, self-supervised pre-training is a possible solution, with contrastive learning dominant for location encoders. Those approaches usually align geographic coordinates with just one additional modality. We propose two multimodal contrastive learning architectures: Multimodal Embedding via Location Tying (\textsc{melt}) and Sequential Alternating Location Training (\textsc{salt}). These architectures expand this framework beyond two modalities by utilising unpaired geospatial data. Both methods are technically viable and match the performance of the strongest two-modality baseline (\textsc{SatCLIP}) across four downstream tasks. However, increasing the number of modalities does not consistently improve performance, suggesting that the chosen location encoder is the main limitation -- the contrastive objective reaches its peak early, regardless of modality diversity or pre-training volume. \textsc{melt} provides more stable training than \textsc{salt} and presents a stronger foundation for future scaling.
\end{abstract}
 \section{INTRODUCTION} 
Diverse large-scale spatial prediction tasks ranging from ecological modelling to human population dynamics are fundamentally constrained by the scarcity of labelled ground-truth observations. While unlabelled geospatial data in different modalities (satellite imagery, geotagged photographs, encyclopaedic text) is available at scale, supervised models cannot exploit it without labels. Earth embedding models close this gap by learning transferable vector representations of geospatial data at specific locations from unlabelled data \cite{c27}. Earth embeddings can be learnt explicitly, by extracting them from raw data such as satellite imagery, or implicitly, by mapping geographic coordinates, e.g., via a location encoder, to an embedding. Those location encoders store the location information within the model weights. Therefore, they are particularly attractive for transfer to arbitrary downstream tasks without modality-specific input at inference time \cite{c6}. By providing rich, learnt representations of geographic locations, these embeddings enable better predictions even when only a limited number of labelled observations are available for a given task.\\
Contrastive learning (CL) is currently the dominant paradigm for pre-training location encoders \cite{c27,c7,c8,c9,c10}. Existing methods pair geographic coordinates with a single modality, which is mostly trained via NT-Xent loss. The two-modality paradigm captures only the pairwise information between a location and a single modality, thereby limiting the entropy that can be extracted from geographic data. While one modality might carry spectral reflectance, others provide distinct semantic and/or visual contexts. By including multiple modalities, the objective should ideally force the encoder to combine these disparate signals into a unified embedding. Thereby attempting to maximise the synergistic information from the multiple views.\\
To the best of our knowledge, no prior work has studied this empirically in the context of implicit earth embeddings. Related work on explicit earth embedding models, such as \textsc{AlphaEarth} \cite{c17}, has incorporated multiple modalities but does so via modality-specific encoders without a shared geographic anchor.\\
This paper addresses this gap by proposing two multimodal CL architectures for training location encoders. These architectures use geographic location as the shared anchor across unpaired modalities, eliminating the need for synchronised multimodal observations. Our contributions are as follows:
\begin{itemize}
  \item \textsc{melt} (Multimodal Embedding via Location Tying): a
    batch-construction approach that jointly trains all modalities
    in each step via a shared contrastive objective.
  \item \textsc{salt} (Sequential Alternating Location Training): an
    epoch-level schedule that alternates the active non-location modality
    while keeping the location encoder permanently active.
  \item An empirical finding that the location encoder itself,
    rather than modality diversity, constitutes the performance ceiling, which can direct future architecture choices.
\end{itemize}
\section{RELATED WORK}
Geodata has characteristics that distinguish it from other types of data. These differences must be explicitly addressed within the model accordingly. One example is spatial autocorrelation, which describes the tendency for nearby locations to share similar attribute values. This aligns with Tobler’s First Law of Geography, which states that near things are more related than distant things. This phenomenon has historically motivated the development of spatial extensions to standard models, e.g. Kriging or spatial regression variants \cite{c1, c2}. Established approaches encode spatial structure through explicit geometric relationships. In contrast, recent work in GeoAI argues for learning such structure directly from data \cite{c3}. This shift motivates the development of earth embedding models as a flexible and learnable alternative.\\
Within the wider discourse of Geospatial Foundation Models, two primary paradigms are discussed \cite{c18}. On the explicit side, models such as \textsc{SatMAE} \cite{c4} or \textsc{RemoteCLIP} \cite{c5} build general-purpose representations from remote sensing imagery, requiring image input at inference time. In contrast, this work focuses on the implicit side, where a location encoder maps raw geographic coordinates $\mathbf{x} \in \mathbb{R}^2$ into a high-dimensional embedding $\mathbf{z} \in \mathbb{R}^d$ that can be consumed by arbitrary downstream models. With that, no modality-specific input is required at inference time. Following \cite{c6}, such an encoder should preserve spatial distance, support inductive generalisation to unseen coordinates, and remain task-agnostic. Those properties make it a uniquely flexible building block for various spatial prediction tasks.\\
As already indicated, location encoders are most often proposed and trained using CL. In this approach, geographic coordinates are paired with a single additional modality. For instance, \textsc{SatCLIP} \cite{c7} aligns global Sentinel-2 satellite patches with coordinates via a spherical-harmonic-based encoder. \textsc{GeoCLIP} \cite{c8} pairs coordinates with geotagged natural images using random Fourier features and hierarchical scale representations. CSP \cite{c9} similarly uses either natural images or remote sensing imagery, whereas \textsc{RANGE} \cite{c10} extends this line of work by proposing an additional processing step for the location encoder.\\
Besides approaches which produce earth embeddings, related ideas incorporate spatial data and follow a similar research direction. For example, \cite{c19} combines spatial image, text and sound data via a combined loss. Both \cite{c20} and \cite{c21} build on the \textsc{ImageBIND} \cite{c22} logic, in which a new embedding space is not created via alignment, but rather additional modalities are connected to an existing space.\\
To the best of our knowledge, no existing work has extended the implicit model setup beyond two modalities. The potential of utilising additional data sources such as text, natural images, and satellite imagery simultaneously has been noted theoretically \cite{c6, c7, c11}, but has not yet been empirically studied. This gap motivates the present work, which proposes and assesses two architectures, \textsc{melt} and \textsc{salt}, for training a location encoder within a multimodal CL framework using location as the tying element across more than two modalities.  
\section{METHODOLOGY} 
\begin{figure*}[!tb]
  \centering
  \def\svgwidth{0.99\textwidth}
  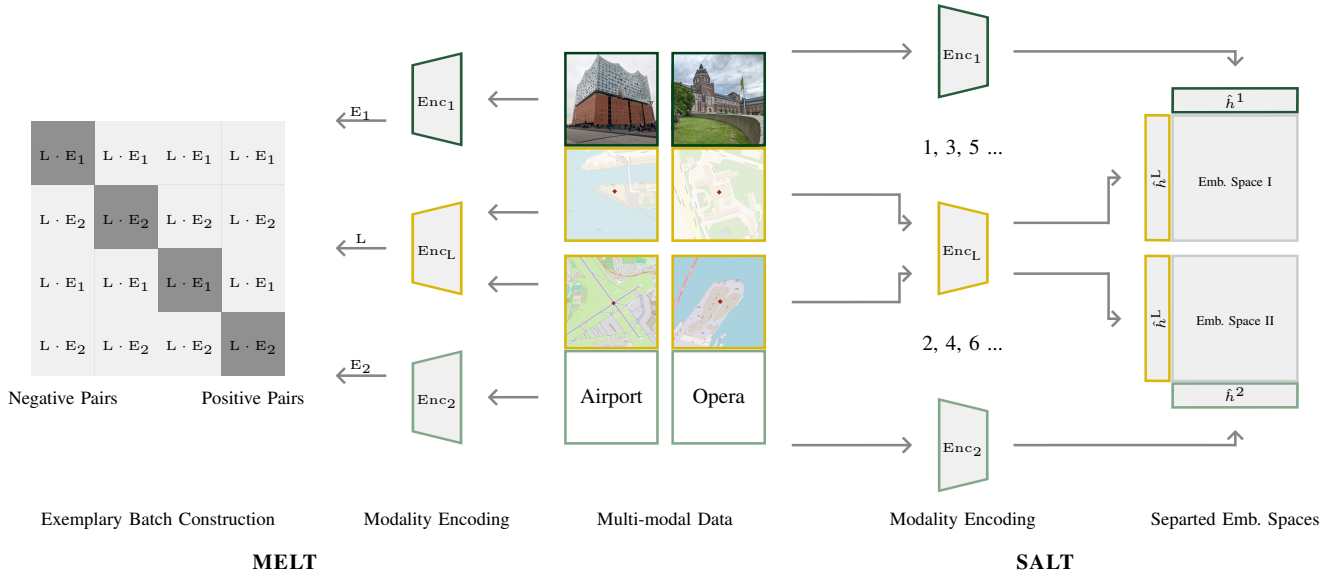
  \caption{Overview of \textsc{melt} and \textsc{salt} approaches. Both start with some multimodal data sources, which is encoded. For \textsc{melt} on the left side, it can be seen that one batch is constructed directly with an equal number of pairs from $\mathrm{Enc}_{\text{1}}$ and $\mathrm{Enc}_{\text{2}}$. For the \textsc{salt} approach, depending on the epoch, always two modalities are paired directly. Indirectly the alignment is conducted via $\mathrm{Enc}_{\text{L}}$.}
  \label{fig:melt_salt}
\end{figure*}
\subsection{Problem Definition}
Let $\mathcal{M} = \{m_0, m_1, \dots, m_{K-1}\}$ denote a set of $K$ data modalities, with each modality $m_k$ corresponding to a particular type of geospatial information. In this setup, $m_0$ is designated as the location modality, which encodes geographic coordinates. The goal is to pre-train a task-agnostic location encoder $h_\theta^{(m_0)}$, parametrised by learnable weights $\theta$, which maps coordinates from a sphere $\mathbb{S}^2$ on a $d$-dimensional continuous representation, \begin{equation}   h_\theta^{(m_0)} : \mathbb{S}^2 \rightarrow \mathbb{R}^d, \quad   \mathbf{x} = (\lambda, \phi) \in \mathbb{S}^2, \end{equation} with $\mathbf{x}$ denoting the geographic location defined by its longitude $\lambda$ and latitude $\phi$. Pre-training aligns $h_\theta^{(m_0)}$ with additional modalities via contrastive learning. For each modality $m \in \mathcal{M}$, a backbone $h_\theta^{(m)}$ and a modality-specific projection head $k_\gamma^{(m)}: \mathbb{R}^d \rightarrow \mathbb{R}^{d'}$ produce the contrastive embedding \cite{c12}: \begin{equation}   z_i^{(m)} = k_\gamma^{(m)}\!\bigl(h_\theta^{(m)}(x_i^{(m)})\bigr), \end{equation} where $x_i^{(m)}$ is the $i$-th input sample of modality $m$, $\gamma$ is the learnable parameters of the projection head, and $z_i^{(m)} \in \mathbb{R}^{d'}$ is the resulting contrastive embedding in a shared $d'$-dimensional space.\\
Another important aspect of our approach concerns the nature of the modality datasets. A key design constraint is that the modality datasets are unpaired: there is no direct one-to-one correspondence between non-location modalities, nor is one assumed. This reflects the practical reality of geospatial data, where different modalities are collected independently at different locations and resolutions. Rather than requiring synchronised multi-modal observations, we exploit geographic location as a natural bridge between modalities: spatial co-location serves as a proxy for semantic relatedness. Concretely, for each sample $i$ from modality $m \neq m_0$, its sole positive counterpart is its co-located location projection $z_i^{(m_0)}$:
\begin{equation}
  \mathcal{P}_i = \bigl\{\bigl(z_i^{(m)},\; z_i^{(m_0)}\bigr)\bigr\}.
\end{equation}
Through this tying mechanism, the location encoder acts as a shared anchor that implicitly aligns all modalities in a common embedding space, without requiring direct cross-modal pairing.
\subsection{\textsc{melt}: Multimodal Embedding via Location Tying}
The location tying construction defined in (3) naturally suggests a joint training regime. In this setup, every non-location sample already has a designated positive (its co-located location projection). All modalities can therefore be trained simultaneously within a single batch, pulling their embeddings towards a common anchor. \textsc{melt} realises this idea directly.\\
\textbf{Batch construction:} In every step of the training process, $N$ samples are drawn from each non-location modality $m \in \mathcal{M}\setminus\{m_0\}$, and each sample is paired with its co-located location embedding. Let $\mathcal{B}^{(m)} = \{z_1^{(m)}, \dots, z_N^{(m)}\}$ denote the resulting set of contrastive embeddings for modality $m$ within a batch. The batch therefore contains $(K-1)\cdot N$ contrastive pairs. For each anchor, the loss contrasts its positive location embedding against all other location embeddings in $\mathcal{B}^{(m_0)}$.\\
\textbf{Objective:} Given the positive pairs $\mathcal{P}_i$ from (3), the training signal is a symmetric NT-Xent loss~\cite{c12} aggregated over all non-location
modalities:
\begin{equation}
  \mathcal{L}_{\text{MELT}} =
    \frac{1}{|\mathcal{M}|-1}
    \sum_{m \in \mathcal{M} \setminus \{m_0\}}
    \mathcal{L}_{\text{NT-Xent}}\!\bigl(
      \mathcal{B}^{(m)},\, \mathcal{B}^{(m_0)}
    \bigr),
\end{equation}
where, for anchor $i$ from modality $m$, the per-sample loss contrasts its positive location embedding against all location embeddings in the batch:
\begin{equation}
  \ell(i) = -\log
    \frac{%
      \exp\!\bigl(\operatorname{sim}(z_i^{(m)},\, z_i^{(m_0)})/\tau\bigr)
    }{%
      \displaystyle\sum_{z_j^{(m_0)} \in \mathcal{B}^{(m_0)}}
      \exp\!\bigl(\operatorname{sim}(z_i^{(m)},\, z_j^{(m_0)})/\tau\bigr)
    },
\end{equation}
where $\ell(i)$ denotes the directional loss for anchor $i$ from modality $m$ towards $m_0$. The symmetric loss is then computed as the average of the directional loss $\ell_{m \rightarrow m_0}$ and $\ell_{m_0 \rightarrow m}$. The cosine similarity is represented as $\operatorname{sim}(\cdot,\cdot)$ and a learnable temperature $\tau > 0$. All encoder parameters $\{\theta^{(m)},\gamma^{(m)}\}_{m \in \mathcal{M}}$ are updated jointly.\\
This results in a single shared embedding space emerging from a single training pass. Alternating updates are not needed. An overview of the architecture is shown in Fig.~\ref{fig:melt_salt}.
\subsection{\textsc{salt:} Sequential Alternating Location Training}
\textsc{salt} uses the same loss and encoder, but does not jointly train all modalities. Instead, it exposes the location encoder to a single non-location modality at a time, alternating modalities at each epoch. A deterministic schedule $\sigma : \mathbb{N} \rightarrow 2^{\mathcal{M}}$ selects the active modality set per epoch $e$:
\begin{equation}
  \sigma(e) = \bigl\{m_0,\; m_{(e \bmod (K-1)) + 1}\bigr\}.
\end{equation}
The constraint $m_0 \in \sigma(e)\;\forall\, e$ ensures the location encoder receives a gradient signal at every epoch. Only the parameters of the currently active modality are updated:
\begin{equation}
    \min_{\theta^{(m)}\!,\,\gamma^{(m)}\!,\, m \in \sigma(e)} 
    \mathcal{L}_{\text{NT-Xent}}( \mathcal{B}^{(m_e)}\!, \mathcal{B}^{(m_0)} ), \,\, m_e = \sigma(e) \setminus \{m_0\}
\end{equation}
This approach allows for adding modalities without architectural change. However, it introduces only indirect inter-modal information flow. Modalities $m_a$ and $m_b$ are never directly compared. They are only linked via a shared training signal with $m_0$. Fig.~\ref{fig:melt_salt} provides an illustration.
\subsection{Encoders and Training Setup}
The location encoder follows \cite{c23} and uses spherical harmonic basis functions up to tenth degree as positional encoding, combined with a SIREN MLP \cite{c13} (\num{2} blocks, \num{512} hidden dimension, \num{445}K parameters).\\
For the non-location modalities, pre-trained backbones are kept frozen, with only a trainable linear adapter added on top.  Satellite imagery uses a MoCo-pre-trained ResNet-18 (${\sim}$\num{131}K trainable parameters) \cite{c32}, natural images a DINO-pre-trained Vision Transformer (${\sim}$\num{99}K trainable parameters) \cite{c31}, and text a BGE encoder \cite{c14} (${\sim}$\num{165}K trainable parameters).  All projection heads map to $d'{=}256$. Training uses AdamW (lr\,$=$\,\num{1e-4}, weight decay\,$=$\,\num{1e-2}) over \num{50} epochs with an accumulated batch size of \num{2048} and a learnable temperature initialised at $\tau{=}0.07$.
\section{EXPERIMENTAL SETUP}
\subsection{Pre-training Data}
Three non-location modalities are used for pre-training. Each modality targets ${\sim}$ \num{100}K globally sampled land mass locations, using a \num{90}/\num{10} train/validation split.\\
\textbf{Satellite imagery:} The S2-100K dataset from \cite{c7} provides 86\,648 Sentinel-2 L2A patches (12 bands, \SI{10}{\meter} resolution, $256{\times}256$ pixels), sampled approximately uniformly across global land mass with ${<}20\%$ cloud cover.\\
\textbf{Natural images:} \num{100000} geotagged images are drawn from the MP-16 subset \cite{c24} of YFCC100M \cite{c25}, comprising user-uploaded Flickr imagery with diverse semantic content.\\
\textbf{Text:} \num{100000} English Wikipedia articles with geographic coordinates (Wikidata property P625) are concatenated per spatial cell, each cell covering ${\approx}\SI{2560}{\meter}{\times}\SI{2560}{\meter}$.\\
Since both the natural-image and text datasets show strong spatial concentration towards western countries and urban areas, kernel density inverse weighting is applied during sampling to reduce geographic bias. Geographic coordinates are augmented via random jitter ($r{=}\SI{0.01}{\degree}$). Images also receive random flips, cropping, brightness/contrast adjustments, and Gaussian blur.
\subsection{Pre-training Variants}
Each architecture (\textsc{melt}, \textsc{salt}) comprises twelve model configurations, defined by the combination of modalities (\textsc{s}\,=\,Satellite, \textsc{t}\,=\,Text, \textsc{n}\,=\,Natural image) and the data regime. Dual-modality baselines (\textsc{SatCLIP}, \textsc{NatCLIP}, \textsc{TextCLIP}) are trained under the same protocol to isolate the effect of additional modalities.\\
Three data regimes are used to control for pre-training volume. In the \textbf{-10} regime, \num{10000} training pairs are distributed equally across active modalities, simulating limited data availability. In the \textbf{-50} regime, 100\,000 pairs are provided in total, creating a high-data comparison at constant volume. The \textbf{-100} regime uses all available data per modality: up to \num{100000} for dual-modality models, up to \num{200000} for three-modality, and ${\sim}$\num{261000} for the full four-modality setup (bounded by the \num{86648} satellite patches, since \textsc{melt} requires equal modality counts per batch).
\subsection{Downstream Tasks}
All models are evaluated on four geographically global tasks: \textbf{Elevation} (metres above sea level, abbr. Elev., \cite{c29}), \textbf{population density} (log-transformed, $\log(1{+}x)$, abbr. Pop., GPW v4 \cite{c26}), \textbf{country codes} (\num{185} classes, abbr. Ctry., \cite{c28}), and \textbf{biomes} (\num{14} classes, abbr. Bio., \cite{c15}).\\
Regression is evaluated with $R^2$, and classification with top-1 accuracy. Downstream training sets use sizes $n \in \{3, 4, \dots, 10\} (= 243, 1\,024, \dots, 100\,000$), uniformly sampled from the land mask via rejection sampling, with \num{20}\% held out for validation. The test set always contains \num{100000} independent samples.\\
The frozen location embedding is fed to a downstream MLP. The hyperparameters are tuned via random search (10 trials, \num{100} epochs) over hidden size $\in \{32, 64, 128, 256\}$, layers $\in \{1,2,3,4\}$, learning rate $\in \{10^{-4}, 10^{-3}, 10^{-2}\}$, and weight decay $\in \{0, 10^{-2}, 10^{-4}\}$. The best configuration is retrained for 500 epochs over ten random initialisations. We report the mean of the metrics and its standard deviation.
\subsection{Baselines}
We compare against six external models in two categories. As alternative location encoders, we use \textsc{SatCLIP-Org} \cite{c7} (the published \num{500}-epoch weights), \textsc{GeoCLIP} \cite{c8}, \textsc{CSP-iNat} and \textsc{CSP-fMoW} \cite{c9}, and \textsc{Range} \cite{c10}. It should be noted that these are pre-trained on different datasets.\\
As non-learned baselines, we include \textsc{Mosaiks}, which combine pre-computed random convolutional features from satellite imagery (\SI{0.1}{\degree} grid, \num{4000}-dim) \cite{c16}. We also include \textsc{Identity}, which passes raw normalised coordinates directly to the downstream MLP.
\section{RESULTS}
\begin{figure}
    \centering
    \resizebox{0.99\columnwidth}{!}{%
    \input{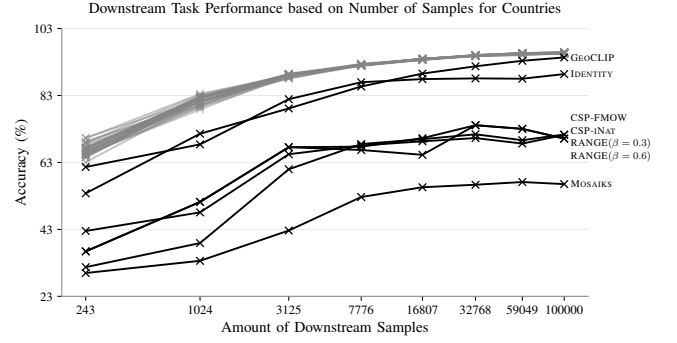}
    }
    \caption{Accuracy for different amounts for the countries' downstream task. Black lines depict all the different baseline models, and the combined variants of \textsc{salt} and \textsc{melt} are shown in grey.}
    \label{fig:base_all_data_amount_countries}
\end{figure}
\subsection{Quantitative Results}
\begin{table*}[t]
\renewcommand{\arraystretch}{1.2}
\centering
\small
\caption{Selected Results: Baselines vs. Proposed MELT/SALT Variants. The \textbf{best} and \underline{second-best} results are highlighted.}
\resizebox{0.99\textwidth}{!}{%
\begin{tabular}{l rrrr rrrr rrrr}
\toprule
& \multicolumn{4}{c}{\textbf{243 samples}} & \multicolumn{4}{c}{\textbf{1\,024 samples}} & \multicolumn{4}{c}{\textbf{3\,125 samples}} \\
\cmidrule(lr){2-5} \cmidrule(lr){6-9} \cmidrule(lr){10-13}
Model & Elev. & Ctry. & Pop. & Bio. & Elev. & Ctry. & Pop. & Bio. & Elev. & Ctry. & Pop. & Bio.\\
\midrule
\textit{Baselines} & & & & & & & & & & & & \\
\textsc{SatCLIP-Org}   & \textbf{.57}{\color{gray}{\tiny±.03}} & 65.4{\color{gray}{\tiny±0.9}} & .47{\color{gray}{\tiny±.02}} & 66.8{\color{gray}{\tiny±0.7}} & \textbf{.74}{\color{gray}{\tiny±.00}} & 79.5{\color{gray}{\tiny±0.5}} & \underline{.62}{\color{gray}{\tiny±.01}} & \textbf{78.1}{\color{gray}{\tiny±0.1}} & \textbf{.80}{\color{gray}{\tiny±.01}} & 88.4{\color{gray}{\tiny±0.1}} & \textbf{.68}{\color{gray}{\tiny±.01}} & \underline{82.3}{\color{gray}{\tiny±0.4}} \\
\textsc{GeoCLIP}       & .34{\color{gray}{\tiny±.00}} & 53.8{\color{gray}{\tiny±1.3}} & .47{\color{gray}{\tiny±.01}} & 62.4{\color{gray}{\tiny±1.3}} & .54{\color{gray}{\tiny±.01}} & 71.6{\color{gray}{\tiny±0.7}} & .48{\color{gray}{\tiny±.01}} & 69.4{\color{gray}{\tiny±0.5}} & .62{\color{gray}{\tiny±.01}} & 79.1{\color{gray}{\tiny±0.5}} & .56{\color{gray}{\tiny±.01}} & 75.5{\color{gray}{\tiny±0.2}} \\
\textsc{Identity}      & .18{\color{gray}{\tiny±.03}} & 61.6{\color{gray}{\tiny±1.7}} & .39{\color{gray}{\tiny±.03}} & 63.1{\color{gray}{\tiny±1.6}} & .23{\color{gray}{\tiny±.13}} & 68.4{\color{gray}{\tiny±1.4}} & .59{\color{gray}{\tiny±.02}} & 72.0{\color{gray}{\tiny±1.0}} & .64{\color{gray}{\tiny±.04}} & 81.9{\color{gray}{\tiny±0.5}} & .61{\color{gray}{\tiny±.03}} & 78.5{\color{gray}{\tiny±0.3}} \\
\textsc{Range}         & .37{\color{gray}{\tiny±.03}} & 36.4{\color{gray}{\tiny±1.7}} & .41{\color{gray}{\tiny±.12}} & 55.7{\color{gray}{\tiny±1.9}} & .72{\color{gray}{\tiny±.12}} & 51.2{\color{gray}{\tiny±2.7}} & .47{\color{gray}{\tiny±.32}} & 67.2{\color{gray}{\tiny±5.2}} & .71{\color{gray}{\tiny±.11}} & 67.6{\color{gray}{\tiny±8.0}} & .64{\color{gray}{\tiny±.02}} & 71.0{\color{gray}{\tiny±11.7}} \\
\textsc{CSP-FMOW}         & .22{\color{gray}{\tiny±.04}} & 42.5{\color{gray}{\tiny±3.1}} & .26{\color{gray}{\tiny±.05}} & 58.6{\color{gray}{\tiny±0.5}} & .34{\color{gray}{\tiny±.01}} & 48.1{\color{gray}{\tiny±2.8}} & .37{\color{gray}{\tiny±.04}} & 62.1{\color{gray}{\tiny±1.4}} & .39{\color{gray}{\tiny±.03}} & 65.5{\color{gray}{\tiny±1.8}} & .39{\color{gray}{\tiny±.06}} & 66.2{\color{gray}{\tiny±0.9}} \\
\textsc{CSP-iNat}         & .01{\color{gray}{\tiny±.06}} & 31.7{\color{gray}{\tiny±3.2}} & .02{\color{gray}{\tiny±.03}} &42.5{\color{gray}{\tiny±3.1}} & .19{\color{gray}{\tiny±.03}} & 38.9{\color{gray}{\tiny±0.9}} & .35{\color{gray}{\tiny±.03}} & 57.3{\color{gray}{\tiny±1.3}} & .20{\color{gray}{\tiny±.02}} & 60.9{\color{gray}{\tiny±0.5}} & .41{\color{gray}{\tiny±.02}} & 63.5{\color{gray}{\tiny±0.5}} \\ 
\textsc{Mosaiks}       & .04{\color{gray}{\tiny±.09}} & 30.0{\color{gray}{\tiny±0.7}} & .30{\color{gray}{\tiny±.03}} & 53.0{\color{gray}{\tiny±3.2}} & .34{\color{gray}{\tiny±.03}} & 33.6{\color{gray}{\tiny±1.3}} & .33{\color{gray}{\tiny±.23}} & 65.3{\color{gray}{\tiny±0.5}} & .41{\color{gray}{\tiny±.01}} & 42.7{\color{gray}{\tiny±1.4}} & .47{\color{gray}{\tiny±.05}} & 68.7{\color{gray}{\tiny±0.3}} \\
\midrule
\textit{Dual Modality} & & & & & & & & & & & & \\
\textsc{SatCLIP-100}   & .48{\color{gray}{\tiny±.01}} & 65.1{\color{gray}{\tiny±0.9}} & \textbf{.57}{\color{gray}{\tiny±.00}} & 69.1{\color{gray}{\tiny±0.3}} & .71{\color{gray}{\tiny±.01}} & 81.2{\color{gray}{\tiny±0.4}} & .60{\color{gray}{\tiny±.01}} & 76.9{\color{gray}{\tiny±0.3}} & \underline{.79}{\color{gray}{\tiny±.02}} & 89.0{\color{gray}{\tiny±0.2}} & \underline{.67}{\color{gray}{\tiny±.01}} & \underline{82.3}{\color{gray}{\tiny±0.3}} \\
\textsc{TextCLIP-100}  & .26{\color{gray}{\tiny±.11}} & \textbf{70.2}{\color{gray}{\tiny±0.9}} & .46{\color{gray}{\tiny±.01}} & 67.3{\color{gray}{\tiny±0.6}} & .70{\color{gray}{\tiny±.01}} & \textbf{83.5}{\color{gray}{\tiny±0.2}} & .57{\color{gray}{\tiny±.02}} & 76.2{\color{gray}{\tiny±0.2}} & \underline{.79}{\color{gray}{\tiny±.01}} & 89.1{\color{gray}{\tiny±0.1}} & \textbf{.68}{\color{gray}{\tiny±.02}} & 81.9{\color{gray}{\tiny±0.7}} \\
\textsc{NatCLIP-100}   & .46{\color{gray}{\tiny±.00}} & 64.6{\color{gray}{\tiny±0.6}} & .32{\color{gray}{\tiny±.11}} & 66.7{\color{gray}{\tiny±0.6}} & .68{\color{gray}{\tiny±.01}} & 82.5{\color{gray}{\tiny±0.2}} & .60{\color{gray}{\tiny±.01}} & 75.9{\color{gray}{\tiny±0.3}} & .78{\color{gray}{\tiny±.01}} & 89.2{\color{gray}{\tiny±0.2}} & .66{\color{gray}{\tiny±.01}} & 81.8{\color{gray}{\tiny±0.3}} \\
\midrule
\textsc{melt} \textit{(Ours)} & & & & & & & & & & & & \\
STN-10      & \underline{.56}{\color{gray}{\tiny±.01}} & 67.4{\color{gray}{\tiny±0.8}} & \textbf{.57}{\color{gray}{\tiny±.01}} & 67.5{\color{gray}{\tiny±0.6}} & \underline{.73}{\color{gray}{\tiny±.00}} & 81.1{\color{gray}{\tiny±0.6}} & \textbf{.63}{\color{gray}{\tiny±.01}} & 77.2{\color{gray}{\tiny±0.2}} & \textbf{.80}{\color{gray}{\tiny±.00}} & \underline{89.3}{\color{gray}{\tiny±0.2}} & .66{\color{gray}{\tiny±.02}} & \underline{82.3}{\color{gray}{\tiny±0.4}} \\
STN-50      & .44{\color{gray}{\tiny±.11}} & 67.2{\color{gray}{\tiny±1.5}} & \textbf{.57}{\color{gray}{\tiny±.00}} & 69.6{\color{gray}{\tiny±0.5}} & .72{\color{gray}{\tiny±.01}} & 82.7{\color{gray}{\tiny±0.3}} & .58{\color{gray}{\tiny±.02}} & 76.5{\color{gray}{\tiny±0.3}} & \textbf{.80}{\color{gray}{\tiny±.01}} & 88.6{\color{gray}{\tiny±0.2}} & \underline{.67}{\color{gray}{\tiny±.01}} & 82.2{\color{gray}{\tiny±0.4}} \\
STN-100     & \underline{.56}{\color{gray}{\tiny±.01}} & 67.4{\color{gray}{\tiny±0.8}} & \textbf{.57}{\color{gray}{\tiny±.01}} & 69.2{\color{gray}{\tiny±0.2}} & \underline{.73}{\color{gray}{\tiny±.00}} & 81.1{\color{gray}{\tiny±0.6}} & \textbf{.63}{\color{gray}{\tiny±.01}} & 77.2{\color{gray}{\tiny±0.2}} & \textbf{.80}{\color{gray}{\tiny±.00}} & \underline{89.3}{\color{gray}{\tiny±0.2}} & .66{\color{gray}{\tiny±.02}} & \underline{82.3}{\color{gray}{\tiny±0.2}} \\
\textsc{TN-100}         & .51{\color{gray}{\tiny±.04}} & 68.8{\color{gray}{\tiny±0.4}} & .37{\color{gray}{\tiny±.04}} &65.5{\color{gray}{\tiny±0.5}} &
.71{\color{gray}{\tiny±.01}} & \underline{83.2}{\color{gray}{\tiny±0.2}} & .57{\color{gray}{\tiny±.01}} & 76.1{\color{gray}{\tiny±0.3}} & 
\underline{.79}{\color{gray}{\tiny±.01}} & 89.0{\color{gray}{\tiny±0.1}} & \underline{.67}{\color{gray}{\tiny±.03}} & 81.8{\color{gray}{\tiny±0.2}} \\ 
\textsc{SN-100}         & .53{\color{gray}{\tiny±.02}} & 65.4{\color{gray}{\tiny±0.8}} & .33{\color{gray}{\tiny±.04}} &\textbf{70.1}{\color{gray}{\tiny±0.6}}& 
\underline{.73}{\color{gray}{\tiny±.00}} & 81.3{\color{gray}{\tiny±0.2}} & .61{\color{gray}{\tiny±.01}} & \underline{77.4}{\color{gray}{\tiny±0.2}} & 
\textbf{.80}{\color{gray}{\tiny±.01}} & 88.0{\color{gray}{\tiny±0.2}} & \underline{.67}{\color{gray}{\tiny±.01}} & 81.6{\color{gray}{\tiny±0.7}} \\ 
\textsc{ST-100}         & .35{\color{gray}{\tiny±.06}} & 68.8{\color{gray}{\tiny±0.7}} & .45{\color{gray}{\tiny±.02}} &\underline{69.9}{\color{gray}{\tiny±0.6}} & 
.72{\color{gray}{\tiny±.00}} & 81.3{\color{gray}{\tiny±0.4}} & \textbf{.63}{\color{gray}{\tiny±.00}} & \underline{77.4}{\color{gray}{\tiny±0.2}} & 
\textbf{.80}{\color{gray}{\tiny±.01}} & \textbf{89.6}{\color{gray}{\tiny±0.1}} & .66{\color{gray}{\tiny±.01}} & 82.2{\color{gray}{\tiny±0.7}} \\ 

\midrule
\textsc{salt} \textit{(Ours)} & & & & & & & & & & & & \\
\textsc{STN-10}         & .48{\color{gray}{\tiny±.01}} & 64.5{\color{gray}{\tiny±0.6}} & .32{\color{gray}{\tiny±.03}} &66.5{\color{gray}{\tiny±0.4}} &
.69{\color{gray}{\tiny±.01}} & 82.2{\color{gray}{\tiny±0.2}} & .60{\color{gray}{\tiny±.01}} & 76.7{\color{gray}{\tiny±0.2}} & 
.77{\color{gray}{\tiny±.01}} & 89.2{\color{gray}{\tiny±0.1}} & .66{\color{gray}{\tiny±.02}} & 81.7{\color{gray}{\tiny±0.3}} \\ 
\textsc{STN-50}         & .39{\color{gray}{\tiny±.03}} & 65.3{\color{gray}{\tiny±1.3}} & .47{\color{gray}{\tiny±.01}} &67.9{\color{gray}{\tiny±0.1}} &
.69{\color{gray}{\tiny±.01}} & 78.8{\color{gray}{\tiny±0.8}} & .60{\color{gray}{\tiny±.01}} & 76.8{\color{gray}{\tiny±0.1}} & 
.78{\color{gray}{\tiny±.01}} & 89.0{\color{gray}{\tiny±0.1}} & \underline{.67}{\color{gray}{\tiny±.01}} & 82.2{\color{gray}{\tiny±0.1}} \\ 
STN-100     & .48{\color{gray}{\tiny±.01}} & 65.9{\color{gray}{\tiny±0.7}} & .50{\color{gray}{\tiny±.02}} & 66.6{\color{gray}{\tiny±3.4}} & .69{\color{gray}{\tiny±.00}} & 82.5{\color{gray}{\tiny±0.3}} & \underline{.62}{\color{gray}{\tiny±.01}} & 77.2{\color{gray}{\tiny±0.2}} & .78{\color{gray}{\tiny±.01}} & 89.1{\color{gray}{\tiny±0.1}} & \underline{.67}{\color{gray}{\tiny±.01}} & \underline{82.3}{\color{gray}{\tiny±0.1}} \\
TN-100      & .50{\color{gray}{\tiny±.00}} & \underline{69.3}{\color{gray}{\tiny±0.6}} & .38{\color{gray}{\tiny±.03}} & 66.9{\color{gray}{\tiny±0.4}} & .70{\color{gray}{\tiny±.01}} & 80.7{\color{gray}{\tiny±0.5}} & .59{\color{gray}{\tiny±.02}} & 76.6{\color{gray}{\tiny±0.2}} & \underline{.79}{\color{gray}{\tiny±.01}} & 89.0{\color{gray}{\tiny±0.3}} & \textbf{.68}{\color{gray}{\tiny±.01}} & 81.7{\color{gray}{\tiny±0.3}} \\
SN-100      & \underline{.56}{\color{gray}{\tiny±.01}} & 65.8{\color{gray}{\tiny±0.8}} & .38{\color{gray}{\tiny±.03}} & 67.3{\color{gray}{\tiny±0.7}} & .69{\color{gray}{\tiny±.01}} & 82.3{\color{gray}{\tiny±0.1}} & .57{\color{gray}{\tiny±.03}} & 76.9{\color{gray}{\tiny±0.7}} & \underline{.79}{\color{gray}{\tiny±.01}} & 89.1{\color{gray}{\tiny±0.2}} & .65{\color{gray}{\tiny±.01}} & \textbf{82.5}{\color{gray}{\tiny±0.1}} \\
ST-100      & .45{\color{gray}{\tiny±.06}} & 67.3{\color{gray}{\tiny±1.1}} & \underline{.52}{\color{gray}{\tiny±.01}} & 68.9{\color{gray}{\tiny±0.5}} & .71{\color{gray}{\tiny±.00}} & 82.4{\color{gray}{\tiny±0.3}} & .59{\color{gray}{\tiny±.01}} & 77.2{\color{gray}{\tiny±0.2}} & \underline{.79}{\color{gray}{\tiny±.01}} & 89.2{\color{gray}{\tiny±0.4}} & \underline{.67}{\color{gray}{\tiny±.01}} & 81.9{\color{gray}{\tiny±0.4}} \\
\bottomrule
\label{tab:ex_result}
\end{tabular}
}
\end{table*}
\begin{figure}
    \centering
    \resizebox{0.99\columnwidth}{!}{%
        \input{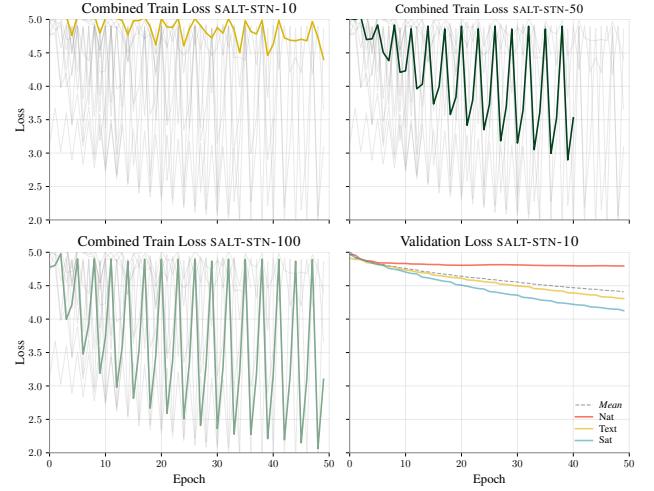}
    }
    \caption{Exemplary training behaviour of \textsc{salt}. Each coloured line highlights the raw combined training loss of a single representative run. All remaining runs are shown in grey and exhibit the similar pattern. The periodic loss spikes visible in every curve coincide with epoch boundaries at which the active modality encoder is switched, partly resulting in
    terminated runs, e.g. \textsc{salt-stn-50}. The validation loss (lower-right inset) is unaffected by this alternation.}
    \label{fig:loss_salt_curve}
\end{figure}
In Tab. \ref{tab:ex_result} we show representative results for all baselines, dual modality models, as well as \textsc{melt} and \textsc{salt} on low training data amounts. For each model the mean metric i.e. $R^2$ for Elev. and Pop. and top-1 accuracy for Ctry. and Bio. plus the standard deviation is shown. Based on the numerical results and supported by additional investigations, three patterns emerge consistently.\\
\textbf{Self-trained vs baselines:} All self-trained models outperform the weaker baselines (\textsc{CSP}, \textsc{Mosaiks}, \textsc{Range}, \textsc{Identity}) by a margin on classification tasks for low and high data regimes. Among baselines, only \textsc{SatCLIP-Org}, which is using the same location encoder, is competitive across the board. An exemplary case can be seen for all downstream data amounts on the countries task in Fig.~\ref{fig:base_all_data_amount_countries}. For both regression tasks, the pattern is similar, but not as strong as for the classification tasks. Some of the self-trained models (\textsc{TextCLIP-100, salt-sn-100}) are surpassed by the baselines under low-data regimes.\\
\textbf{Effect of additional modalities:} Comparing variants of \textsc{melt} and \textsc{salt} across different modalities amount, but the same amount of downstream training data, reveals no consistent gain from the modality. To substantiate this observation statistically, we conduct paired $t$-tests on the population density task, comparing the best- and worst-performing model variants within each configuration across ten random initialisations (Tab.~\ref{tab:global_ttest}). Of all four tested pairs, only \textsc{melt-stn-100} vs. \textsc{melt-sn-50} at \num{243} samples yields a significant difference ($t{=}12.13$, $p{<}0.001$). All remaining pairs are non-significant ($p \geq 0.138$), confirming that performance differences are attributable to random initialisation rather than model design.\\
\textbf{Effect of pre-training data volume:} Scaling pre-training data likewise produces no reliable improvement. Within the \textsc{melt-stn} group, downstream elevation $R^2$ at 243 samples varies non-monotonically across the \textsc{-10, -50, -100} variants. At \num{100000} downstream samples, virtually all self-trained models converge to within one standard deviation of each other and of \textsc{SatCLIP-Org}. These two null results -- no gain from modality diversity, no gain from pre-training volume -- jointly indicate that the location encoder, rather than input amount and richness, constitutes the effective performance ceiling.
\subsection{Qualitative Results}
A first qualitative difference between the two architectures emerges from pre-training dynamics. Fig.~\ref{fig:loss_salt_curve} shows training and validation loss curves for an exemplary \textsc{salt} run. As is clearly visible in the training behaviour, highlighted \textsc{salt} variants exhibit characteristic loss spikes at epoch boundaries where the active encoder switches. This is accompanied by early terminations for some model variants. Such patterns were not observed in the joint batch construction of \textsc{melt}. Despite this instability, \textsc{salt} models achieve final downstream metrics comparable to \textsc{melt}.\\
\begin{figure*}[!tb]
  \centering
  \small
  \def\svgwidth{0.99\textwidth}
  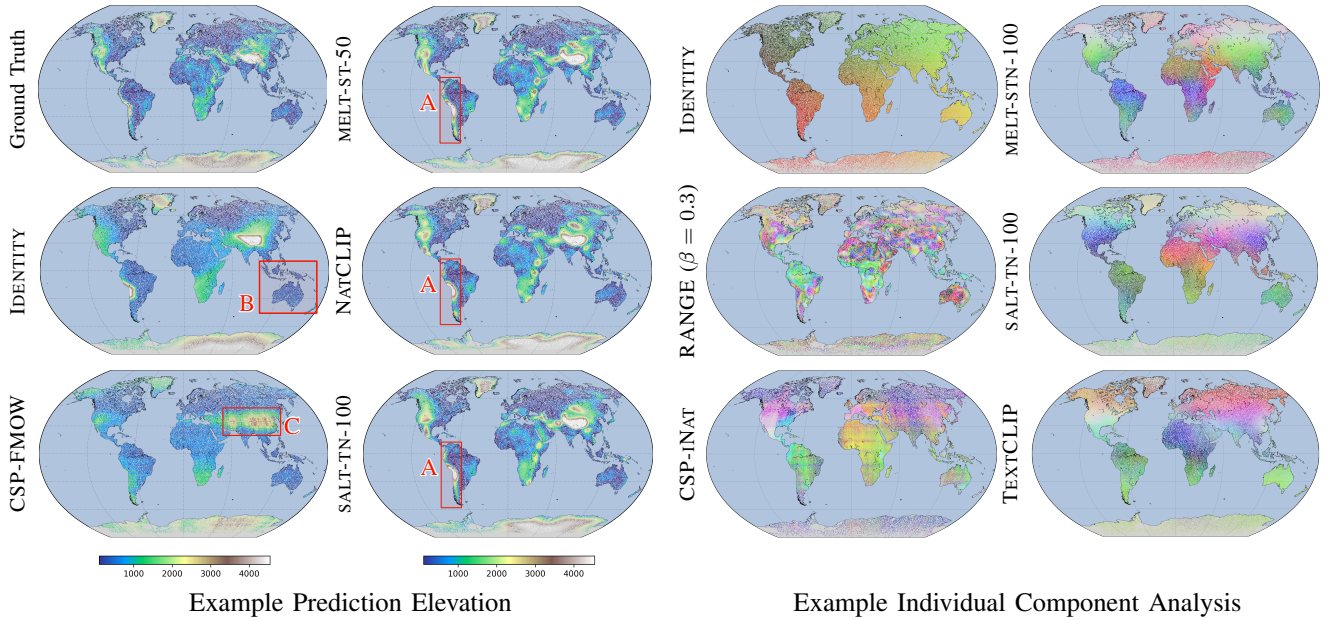
  \caption{The left side shows prediction maps for the elevation task with \num{3125} training samples. From the example of the Andes, it can be observed that the self-trained models produce spatially coherent predictions (A), whereas the baselines show either uniform regions (B) or excessive granularity (C).\\The right side shows ICA embedding visualisations (RGB). Similar structures to those in the prediction map can also be visualised even before, based on the embeddings.}
  \label{fig:pred_maps}
\end{figure*}
The exemplary prediction maps for the elevation task (Fig.~\ref{fig:pred_maps}) show a consistent structural contrast between self-trained and reference models. \textsc{melt} and \textsc{salt} variants produce spatially coherent predictions with large, contiguous areas. In contrast, the different baseline models tend towards either coarse or fine spatial granularity. Notably, prediction patterns appear visually indistinguishable across modality combinations within the self-trained group. This may reflect the quantitative convergence seen above.\\
Independent component analysis (ICA) can be used to visualise the pre-trained embeddings (Fig.~\ref{fig:pred_maps}). This technique helps verify the findings described above in another way. The self-trained encoders produce geographically consistent colour gradients. Ecologically similar regions, e.g. the tropical forests in South America and Central Africa, tend to share similar embedding colours. In contrast, baseline embeddings show grid artefacts (\textsc{CSP}), equatorial discontinuities (\textsc{Identity}), or high variance over regions (\textsc{Range}, \textsc{Mosaiks}).\\
\section{DISCUSSION}
\textbf{Why do additional modalities not improve downstream performances?:} The synergistic information hypothesised in Sec.~I, that knowledge arises only from the combination of multiple views, appears inaccessible under the present setup. The location encoder, constrained by its fixed architecture and low-dimensional coordinate input, appears to saturate early in training: once it has learnt a stable coordinate-to-region mapping, additional modalities provide no gradient signal that the encoder can exploit. In other words, the contrastive objective collapses to a low-complexity alignment task not because the modalities lack complementary information, but because the encoder cannot represent the finer-grained distinctions that such information would require. The disconnect between pre-training loss and downstream quality is further supported through statistical analysis in Tab.~\ref{tab:global_ttest}. Of the four examples, only one shows a significant difference. This verifies that the performance changes are a result of random initialisation and not the modality design. Linear probing offers one diagnostic to confirm this hypothesis. Since a single linear layer yields stagnant performance among all modality combinations, the bottleneck is unambiguously located in the encoder. Rather than in the downstream model's ability to compensate for capacity.\\
\textbf{\textsc{melt} vs \textsc{salt}:} Despite reaching comparable final downstream metrics, the two architectures differ substantially in their training behaviour. Simultaneous batch construction in \textsc{melt} provides a constant, balanced gradient signal to all encoders at every step. This produces smooth loss curves and no early terminations across all twelve variants. \textsc{salt}'s epoch-level switching introduces representational discontinuities at each encoder transition: the location encoder must re-adapt to a new modality distribution at the start of every epoch cycle, leading to the characteristic loss spikes observed in Fig.~\ref{fig:loss_salt_curve} and to premature termination.\\
Despite its instability, \textsc{salt} reaches similar downstream performance. This is further evidence for the bottleneck interpretation. If the alignment task were high in complexity, the representational disruptions caused by encoder switching should produce embeddings that are measurably worse. The observation that the location encoder recovers within a few epochs suggests the target solution is low-dimensional and easy to re-attain — but at the cost of additional computational overhead and reduced reproducibility. \textsc{melt} is therefore the preferred foundation for scaling to more modalities or longer pre-training schedules.\\
\textbf{Limitations:} Four constraints bound the generalisability of the results and should therefore be considered as highly relevant future experiments to validate the findings. First, a fixed one-size-fits-all hyper-parameter strategy (batch size \num{2048}, learning rate \num{1e-4}, \num{50} epochs) was applied across all variants, whereby per-architecture tuning would likely sharpen quantitative differences between \textsc{melt} and \textsc{salt}, and between modality combinations. Especially, the batch size directly impacts the number of negative pairs and therefore creates a limitation on the possible contrast given.\\
Second, only a single location encoder architecture was tested. Thus, the bottleneck hypothesis rests on inductive evidence from a single encoder family and cannot be confirmed without ablating alternative encoders, such as \textsc{GeoCLIP} or \textsc{Range}.\\
Third, the evaluation is restricted to four global tasks. Geographically stratified or few-shot evaluations may reveal advantages of richer multimodal embeddings that the current protocol does not detect. Moreover, deeper embedding inspection methods, such as intrinsic dimensionality \cite{c18}, could clarify the (location) encoders' capacity limits.\\
Lastly, the choice of the NT-Xent objective can be seen as a limiting factor. It has already been argued in this section that the objective leads to a low-complexity alignment task. Whilst the NT-Xent loss is commonly used, its effectiveness must be confirmed using alternative loss functions. At this point, reference can also be made to loss functions in the LeJEPA direction \cite{c33}. Unlike NT-Xent, which shapes the embedding distribution only implicitly through its negative-sampling term, LeJEPA \cite{c33} imposes an explicit, maximum-entropy target distribution and is shown to be minimax-optimal with respect to downstream prediction risk across a broad class of probes. This is expected for the transfer behaviour of a (multimodal) foundation model. Their suitability for geodata has yet to be demonstrated.

\section{CONCLUSION}
We proposed \textsc{melt} and \textsc{salt} as multimodal contrastive learning frameworks for the location encoder, extending beyond two modalities. Using location as the tying element for unpaired geospatial data. Both are technically viable and match the strongest two-modality baselines \textsc{SatCLIP-Org} across four downstream tasks.\\ 
Our main finding is that the location encoder, not modality count or data volume, limits the performance. Neither adding modalities nor scaling pre-training data consistently improved downstream performance beyond that of a two-modality baseline. This points to two complementary limitations. Firstly, the location encoder itself limits the representational potential to create richer multimodal signals. Second, the training objective may be ill-suited for this setting: by pulling all modality embeddings towards a single shared location anchor. Thereby, encouraging the encoder to learn what modalities have in common, but discarding precisely those modality-specific features that would distinguish, for instance, spectral land-cover signatures from semantic textual context. Alternative objectives that preserve modality-specific structure may be necessary to unlock the complementary information provided by additional modalities.\\
Thus, the focus of the upcoming work should be on both. First, alternative encoder architectures that raise the representational ceiling should be tested and second, alignment objectives that move beyond global instance-level similarity should be explored. Beyond these two principal directions, the further constraints discussed in our limitations (hyper-parameter tuning, broader evaluation) equally warrant attention. Given its training reliability, \textsc{melt} provides the stronger practical foundation for these investigations.

\section*{APPENDIX}
\begin{table}[!ht]
\caption{Paired $t$-test results on population density: best vs.\ worst
         model variant per configuration, ten initialisations each.}
\label{tab:global_ttest}
\centering
\scriptsize
\begin{tabularx}{0.99\columnwidth}{@{}l XX r r r@{}}
\toprule
\textbf{$N$} & \textbf{Best} & \textbf{Worst}
  & \textbf{$t$} & \textbf{$p$} & \textbf{$\bar{\Delta}$} \\
\midrule
100K & \textsc{salt-tn-10}   & \textsc{salt-stn-10}
  & $-0.09$ & 0.933             & $-0.001$ \\
100K & \textsc{melt-stn-10}  & \textsc{melt-tn-100}
  & $1.63$  & 0.138             & $0.017$  \\
243  & \textsc{salt-stn-100} & \textsc{salt-st-100}
  & $-0.33$ & 0.750             & $-0.002$ \\
243  & \textsc{melt-stn-100} & \textsc{melt-sn-50}
  & $12.13$ & $\mathbf{{<}\,0.001}$ & $0.129$  \\
\bottomrule
\end{tabularx}
\end{table}

\end{document}